\documentclass{article}

\usepackage{microtype}
\usepackage{graphicx}
\usepackage{subfigure}
\usepackage{booktabs} 

\usepackage{hyperref}

\usepackage[accepted]{icml2024}

\usepackage{amsmath}
\usepackage{amssymb}
\usepackage{mathtools}
\usepackage{amsthm}

\usepackage[capitalize,noabbrev]{cleveref}

\theoremstyle{plain}

\theoremstyle{definition}

\theoremstyle{remark}

\usepackage{booktabs, multirow} 
\usepackage{soul}
\usepackage{ulem}
\newcommand{\zzyu}[1]{\textcolor{black}{#1}}

\usepackage[textsize=tiny]{todonotes}

\icmltitlerunning{Unveiling and Harnessing Hidden Attention Sinks}

\begin{document}

\twocolumn[

\icmltitle{Unveiling and Harnessing Hidden Attention Sinks: Enhancing Large Language Models without Training through Attention Calibration}



\icmlsetsymbol{equal}{*}

\begin{icmlauthorlist}
\icmlauthor{Zhongzhi Yu}{equal,1}
\icmlauthor{Zheng Wang}{equal,1}
\icmlauthor{Yonggan Fu}{1}
\icmlauthor{Huihong Shi}{1}
\icmlauthor{Khalid Shaikh}{1}
\icmlauthor{Yingyan (Celine) Lin}{1}
\end{icmlauthorlist}

\icmlaffiliation{1}{Georgia Institute of Technology}

\icmlcorrespondingauthor{Yingyan (Celine) Lin}{celine.lin@gatech.edu}

\icmlkeywords{Large Language Models, Attention Mechanism, Attention Sink}

\vskip 0.3in
]



\printAffiliationsAndNotice{\icmlEqualContribution} 

\vspace{-2em}
\begin{abstract} 
\vspace{-1em}

Attention is a fundamental component behind the remarkable achievements of large language models (LLMs). However, our current understanding of the attention mechanism, especially regarding how attention distributions are established, remains limited. Inspired by recent studies that explore the presence of attention sink in the initial token, which receives disproportionately large attention scores despite their lack of semantic importance, this work delves deeper into this phenomenon. We aim to provide a more profound understanding of the existence of attention sinks within LLMs and to uncover ways to enhance the achievable accuracy of LLMs by directly optimizing the attention distributions, \textit{without} the need for weight finetuning. Specifically, this work begins with comprehensive visualizations of the attention distributions in LLMs during inference across various inputs and tasks. Based on these visualizations, to the best of our knowledge, we are the first to discover that (1) attention sinks occur not only at the start of sequences but also within later tokens of the input, and (2) not all attention sinks have a positive impact on the achievable accuracy of LLMs. Building upon our findings, we propose a training-free Attention Calibration Technique (ACT) that automatically optimizes the attention distributions on the fly during inference in an input-adaptive manner. \zzyu{Extensive experiments validate that ACT consistently enhances the accuracy of various LLMs across different applications. Specifically, ACT achieves an average improvement of up to 7.30\% in accuracy across different datasets when applied to Llama-30B. 
Our code is available at \url{https://github.com/GATECH-EIC/ACT}.}

\end{abstract}
\vspace{-2em}
\section{Introduction}
\label{sec:intro}
\vspace{-0.3em}

In recent days, large language models (LLMs) have garnered significant attention due to their impressive performance across a wide range of tasks~\cite{touvron2023llama,touvron2023llama2,chatgpt,bard,10323953,gpt4}. One of the key components contributing to the remarkable performance of LLMs is the attention mechanism, which effectively identifies relationships among tokens in a sequence. This ability enables LLMs to comprehend intricate contexts and details, greatly enhancing their capacity to process and generate text that closely resembles human language~\cite{vaswani2017attention,gpt1}. However, despite the immense potential of the attention mechanism, our current understanding of how attention distributions are established and their relationship to the achievable performance of LLMs remains inadequately explored.

Along this direction, a pioneering study, StreamLLM~\cite{xiao2023efficient}, has undertaken an initial investigation and improved our understanding of attention distributions by uncovering the existence of attention sinks. 
In particular, they find that the initial token of an input text receives a disproportionately large attention score, despite often lacking semantic significance. This phenomenon arises from the visibility of the initial token to almost all subsequent tokens in autoregressive language modeling, causing them to become the recipients of these ``unnecessary” attention values. Motivated by the impact of attention sinks on attention distributions, we aim to delve deeper into their general existence to gain a better understanding of how they affect LLMs' reasoning and generation capabilities. This, in turn, will inspire new strategies to enhance the achievable accuracy of LLMs. To achieve this goal, we pose the following three intriguing research questions: 
\textbf{Q1:} \textit{Does an attention sink only exist in the initial token?}
\textbf{Q2:} \textit{Will preserving attention sinks always benefit LLMs' accuracy in different scenarios?}
\textbf{Q3:} \textit{Can we enhance LLMs' accuracy by solely manipulating attention sinks without any weight finetuning?}

In our endeavor to address the aforementioned three questions, 
we make the following contributions:
\vspace{-1em}
\begin{itemize}
    \item We conduct comprehensive visualizations of the attention distributions in LLMs across a variety of tasks and inputs. To the best of our knowledge, we are the first to discover that attention sinks manifest not only in the initial token but also within subsequent tokens throughout the input context. Intriguingly, similar to the attention sink observed in the initial token by~\cite{xiao2023efficient}, attention sinks in later tokens also tend to be concentrated on tokens of less semantic importance.
    \vspace{-1.5em}
    \item 
    Excited by the above observation, we further probe into the relationship between attention sinks at different locations and the accuracy of the generated content at those respective locations. Interestingly, we discover that not all attention sinks have a positive impact on maintaining LLMs' performance, which complements the findings in~\cite{xiao2023efficient}.
    \vspace{-0.5em}
    \item 
    Leveraging the findings above, we have developed a training-free Attention Calibration Technique, named ACT, that automatically optimizes attention distributions on the fly during inference in an input-adaptive manner, improving the achievable accuracy of pretrained LLMs on downstream tasks. Additionally, it can even lead to a comparable accuracy as compared to the commonly used in-context learning technique, and further be combined with the latter for boosted accuracy. As such, our ACT has provided an alternative new design knob for LLM enhancement.  
    \vspace{-0.5em} 
    \item Extensive experiments and ablation studies validate that our proposed method can achieve up to a 7.30\% higher accuracy than the vanilla inference baseline across various tasks. Furthermore, ACT is capable of improving LLMs' performance in challenging multi-round conversation tasks. Specifically, applying ACT to different variants of Llama2 boosts the achievable score by up to 0.13 on the challenging MT-Bench dataset. 
\end{itemize}
\vspace{-0.3em}   
\vspace{-0.9em}
\section{Related Works}
\vspace{-0.3em}

\subsection{Large language models}
\vspace{-0.3em}

Transformer-based language models~\cite{vaswani2017attention,devlin2018bert,2020t5,roberts2022t5x} have demonstrated their remarkable ability to effectively extract relationships among tokens from complex input sequences, thanks to the utilization of the attention mechanism in their model architecture. Furthermore, their attention-centric design enables decent scalability~\cite{qin2023scaling,kaplan2020scaling,biderman2023pythia}: as the model size and pretraining dataset scale increase, the performance of transformer-based language models continues to improve. This phenomenon has given rise to the emergence of LLMs.
One of the earliest impressive LLMs is GPT-3~\cite{gpt3}, which showcases remarkable zero-shot and few-shot in-context learning capabilities. This achievement has further fueled the development of various LLMs, such as OPT~\cite{zhang2022opt}, Llama~\cite{touvron2023llama}, Llama2~\cite{touvron2023llama2}, BLOOM~\cite{workshop2022bloom}, GPT-J~\cite{wang2021gpt}, Pythia~\cite{biderman2023pythia}, and GLM~\cite{du2021glm}. 
These models have further pushed the boundaries of deep learning, gradually moving us toward achieving artificial general intelligence.

\vspace{-0.3em}
\subsection{Parameter-efficient tuning}
\vspace{-0.3em}

Despite the promising zero-shot and few-shot capabilities of LLMs, one common approach to achieving strong performance in real-world applications is to finetune pretrained LLMs for downstream tasks. However, the enormous size of LLMs makes 
traditional weight tuning computationally expensive, requiring significant storage and memory overhead.
To address this challenge, various parameter-efficient tuning (PET) methods have been proposed~\cite{hu2021lora,lester2021power,zhang2020side,sung2022lst,yu2023hint,fu2022losses}. Specifically, instead of updating all parameters in the target LLM, PET selectively updates a small set of learnable modules during finetuning~\cite{qi2023pillow,xia2024chain,zhao2024apt,yu2023master,yu2024edge,zhang2023lora,li2023loftq}. While PET methods can reduce computational, storage, and memory overheads, even state-of-the-art (SOTA) PET methods still face challenges in efficiently finetuning LLMs~\cite{dettmers2023qlora}.
Our proposed method is orthogonal to PET: we aim to enhance the performance of LLMs by directly optimizing attention distributions on the fly during inference, eliminating the need for weight finetuning.

\vspace{-0.3em}
\subsection{Observations regarding LLMs' attention}
\vspace{-0.3em}

\begin{figure*}[ht]
    \centering
    \includegraphics[width=0.95\textwidth]{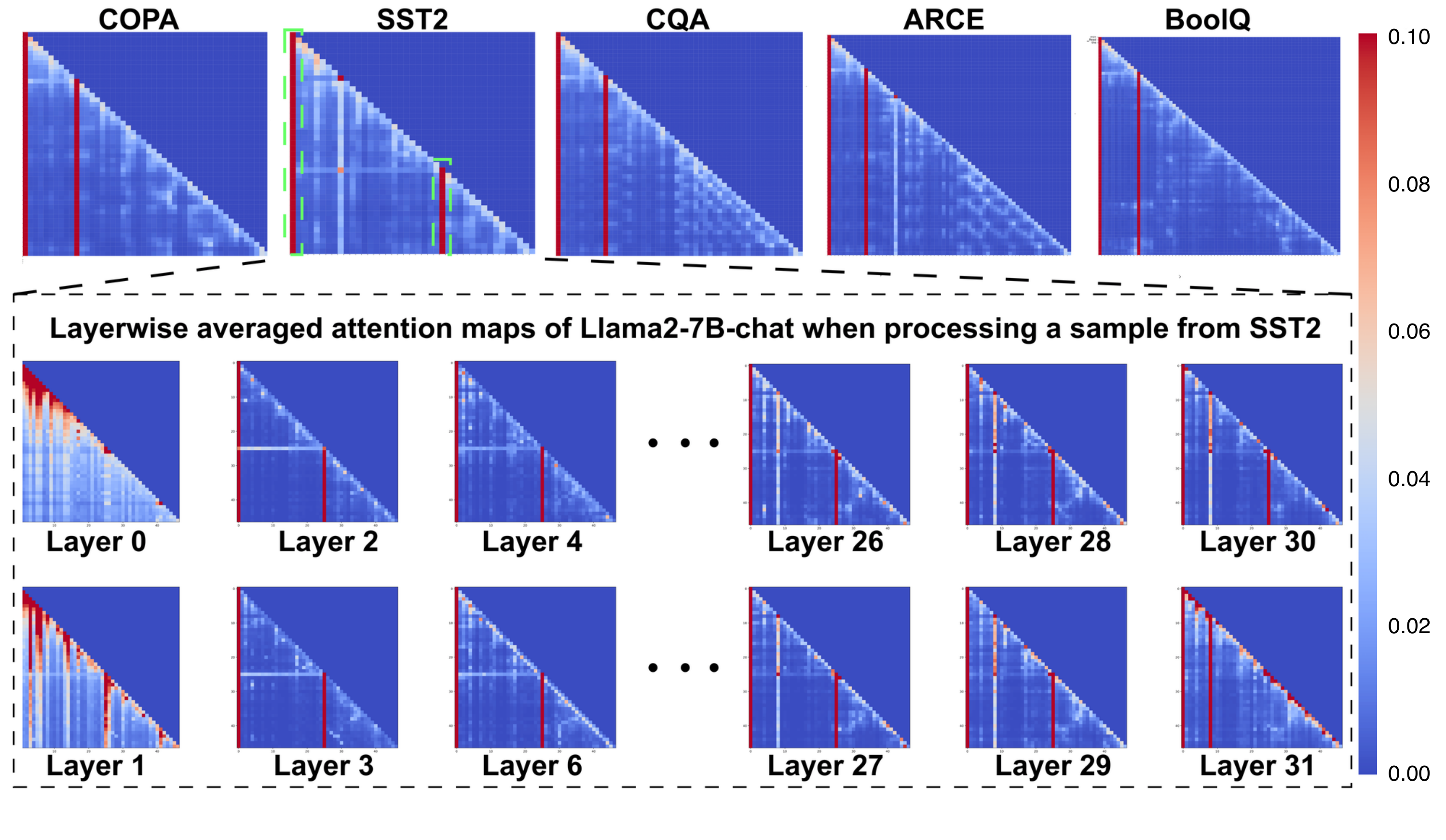}
    \vspace{-2em}
    \caption{Upper: Visualization of the averaged attention maps across all heads and layers of Llama2-7B-chat on different datasets. Lower: Visualization of the averaged attention maps across all heads in each layer when processing a sample from SST2 with Llama2-7B-chat. Identified attention sinks in the averaged attention map from SST2 are bounded with green boxes.}
    \vspace{-1.5em}
    \label{fig:avg_attn}
\end{figure*}

Despite being one of the key components of LLMs, the understanding of the attention mechanism has been slow to evolve compared to the rapid advancement of LLMs themselves. Early works focus on studying attention in small-scale transformers. For instance,~\cite{clark2019does} visualizes specific types of attention patterns in pretrained BERT~\cite{devlin2018bert}, and~\cite{vig2019multiscale} identifies biases and localized relevant attention heads. Additionally,~\cite{sun2020understanding} discovers that the degree of association between a word token and a class label affects their attention score.
However, the exploration of the unique attention distribution in LLMs with larger model sizes and datasets is still in its infancy. Along this trajectory, some pioneering works have made interesting observations related to the attention mechanism in LLMs. For instance,~\cite{kou2023model} finds that the attention distribution in LLMs differs from that in humans, and~\cite{zhang2023tell} observes that increasing the attention score of manually defined tokens at specific heads can improve LLMs' ability to follow instructions. However, determining the relationship between attention distributions and the achievable performance of LLMs, as well as automating the enhancement of LLMs' performance by 
calibrating attention distributions during inference, still remain open challenges.

\vspace{-0.9em}
\section{Preliminaries}
\label{sec:preliminary}
\vspace{-0.3em}

\textbf{LLMs and multi-head attention.}
LLMs~\cite{touvron2023llama, gpt3, gpt4} are typically constructed by stacking $L$ transformer blocks, each comprising a feed-forward network (FFN) and a multi-head attention (MHA) module that captures the pairwise relationships among all $N$ input tokens in the input sequence. Specifically, for a given input $\mathbf{X}^l \in \mathbb{R}^{N \times d}$ to the $l$-th block, the output feature $\mathbf{F}^l_h \in \mathbb{R}^{N \times d}$ generated at head $h$ can be represented as:
\vspace{-0.5em}
\begin{equation}
\begin{split}
    \mathbf{A}^l_h &= {\tt Softmax}\left(\frac{f_Q^l(\mathbf{X}^l) \cdot f_K^l(\mathbf{X}^l)^T}{\sqrt{d_k}}\right), \\ 
    \mathbf{F}_h^l &= \mathbf{A}_h^l \cdot f_V^l(\mathbf{X}^l),
\end{split}
\end{equation}
\vspace{-0.5em}
where $f_Q^l$, $f_K^l$, and $f_V^l$ are projection layers, $d_k = d / h$ is the embedding dimension of each head, and $\mathbf{A}_h^l \in \mathbb{R}^{N \times N}$ is the attention map generated at head $h$. Each element $\mathbf{A}_h^l[i, j]$ represents the relationship between the $i$-th and $j$-th tokens in $\mathbf{X}^l$. The attention score is defined as $a_h^l = [\sum_{j=1}^i \mathbf{A}_h^l[i, j] / i,\ \forall i \in \{1, \cdots, N\}]$, and $a_h^l[i]$ denotes the attention score for the $i$-th token at head $h$, layer $l$.

Next, the features $\mathbf{F}_h^l$ of each head $h$ are combined to generate the output $\mathbf{O}^l$ of MHA by
\vspace{-0.5em}
\begin{equation}
    \mathbf{O}^l = f_O^l({\tt Concat}(\mathbf{F}_1^l, \cdots, \mathbf{F}_h^l)),
\end{equation}
\vspace{-0.5em}
where $f_O^l$ represents a projection layer. In the remainder of this paper, we primarily utilize the distribution of $\mathbf{A}_h^l$ generated by various inputs $\mathbf{X}^l$ for all $h$ and $l$ within the LLM as the key knob to address the three research questions outlined in Sec.~\ref{sec:intro}.


\textbf{StreamLLM and the attention sink.}
StreamLLM~\cite{xiao2023efficient} identifies the presence of an attention sink, which is a token that receives a significantly higher attention score than other tokens but provides limited semantic information. StreamLLM observes that the attention sink only exists in the initial token and suggests always preserving these tokens when processing long input sequences to prevent forgetting.

\vspace{-0.9em}
\section{Unveil and Harness Hidden Attention Sinks}
\label{sec:method}
\vspace{-0.3em}



\textbf{Overview.}
We aim to investigate the general existence of attention sinks and explore their impact on the reasoning and generation process of LLMs. To achieve this goal, we adopt a deductive approach by sequentially addressing three intriguing research questions outlined in Sec.~\ref{sec:intro}:
Firstly, we address \textbf{Q1} to investigate whether attention sinks are limited to the initial token or if they persist in various locations, as discussed in Sec.~\ref{sec:does_exist}.
Secondly, we explore \textbf{Q2} to shed light on the effects of these identified attention sinks on the achievable accuracy of LLMs, as discussed in Sec.~\ref{sec:does_help}.
Finally, building upon the findings gained from \textbf{Q1} and \textbf{Q2}, we address \textbf{Q3} by developing the ACT to enhance the performance of LLMs in a training-free manner during inference, as discussed in Sec.~\ref{sec:how_to_opt}.
Unless otherwise specified, for the remainder of this section, our exploration is based on one of the SOTA LLMs, Llama2-7B-chat~\cite{touvron2023llama2}.

\vspace{-0.3em}
\subsection{\textbf{Q1}: Do attention sinks only exist in the initial token?}
\label{sec:does_exist}
\vspace{-0.3em}

The attention sink has been observed at the initial token of LLMs~\cite{xiao2023efficient}. However, the presence and distribution of attention sinks in later tokens remain an open yet crucial question, especially considering that these tokens contain ample semantic information. Therefore, our objective is to investigate the overall existence of attention sinks that consistently draw significant attention across the entire input sequence.

\begin{table}[t]
    \centering
    \caption{Frequency of tokens appear with significantly higher attention scores.}
    \resizebox{\linewidth}{!}{
    \begin{tabular}{c|ccccc}
        \toprule
        Token name & `$<s>$'& '.'& `$<0x0A>$'& `:'& `Answer'\\ \midrule
        Frequency & 1621135&958992&636902& 65078& 46297\\
        Ratio & 48.2\% & 28.5\% & 18.9\% & 1.9\% & 1.3\%\\
        \midrule
        Token name &  ` '& `Type'& `iment'& `D'& Total\\ \midrule
        Frequency & 21841& 4430& 3896& 2644& 3363296\\
        Ratio & 0.6\% & 0.1\% & 0.1\% & 0.1\% & 100\%\\
        \bottomrule
    \end{tabular}
    }
    \vspace{-2em}
    \label{tab:token_freq}
\end{table}

\begin{figure}
    \centering
    \includegraphics[width=0.9\linewidth]{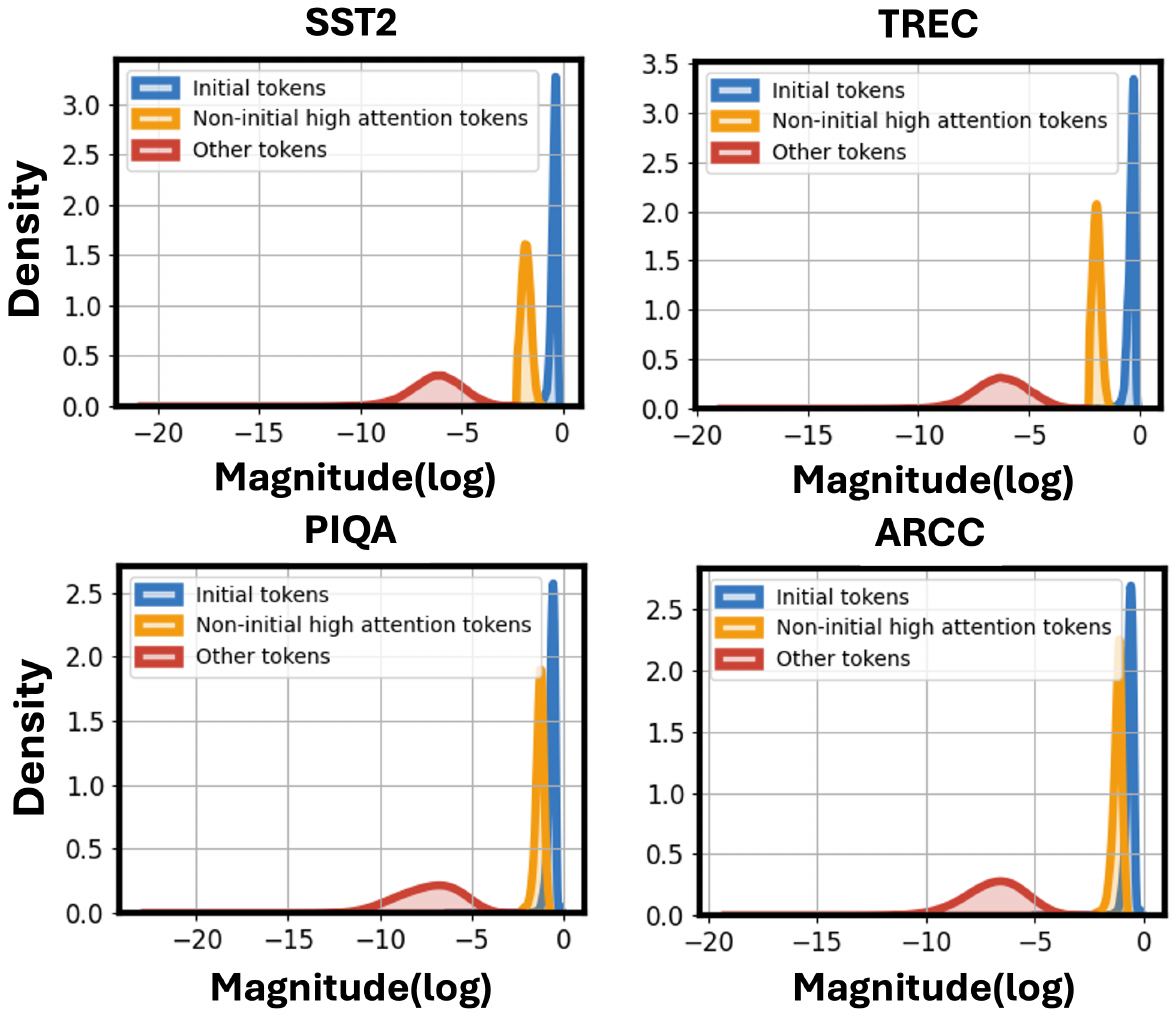}
    \vspace{-1.2em}
    \caption{\zzyu{Attention score distribution of the initial token (i.e., the attention sink observed in StreamLLM~\cite{xiao2023efficient}), non-initial high attention tokens, and other tokens for classification tasks (top) and multiple-choice tasks (bottom).}}
    \vspace{-2em}
    \label{fig:attn_distribution}
\end{figure}

\textbf{Settings.} To address \textbf{Q1}, we first visualize two metrics: (1) the averaged attention maps across all heads and layers, denoted as $(\sum_{h=1}^{H}\sum_{l=1}^{L}\mathbf{A}_h^l)/(H \cdot L)$, on different datasets, and (2) the averaged attention maps of each layer, i.e., $(\sum_{h=1}^{H}\mathbf{A}_h^l)/H)$, when processing a single input sample, as illustrated in Fig.~\ref{fig:avg_attn}. Additional visualizations on various datasets and models can be found in Appendix~\ref{app:avg_attn_vis}.
\zzyu{To generalize these observations across a larger range of datasets, we first visualize the distribution of token-wise attention scores across different datasets to validate the significant gap between high-attention and normal tokens. We further determine that the $i$-th token has a significantly higher attention score if $a_h^l[i] > \alpha / N$ (i.e., more than $\alpha$ times the average attention score) and is considered an attention sink. Specifically, we set $\alpha=5$ based on our upcoming visualization in Fig.~\ref{fig:attn_distribution} unless otherwise specified.}
We summarize the frequency of tokens exhibiting significantly higher attention scores across all samples in a mixed dataset comprising 100 samples collected from each of the 18 datasets mentioned in Sec.~\ref{sec:exp_setting}. 



\textbf{Observations.} 
We can draw the following observations from Fig.~\ref{fig:avg_attn}: \textit{Obs-(1)} \zzyu{several tokens consistently attract significantly higher attention values than other tokens. Moreover, as visualized in Fig.~\ref{fig:attn_distribution}, the distribution of high-attention tokens' attention values has a notable boundary with those of other tokens across different datasets, validating that the difference in attention scores between identified high-attention tokens and other tokens is significant}; \textit{Obs-(2)} as illustrated in Table~\ref{tab:token_freq}, aside from the initial token \textit{$<$S$>$}, which corresponds exactly to the attention sink observed in StreamLLM~\cite{xiao2023efficient}, there also exist a non-trivial number of other attention sinks that contain limited semantic information (e.g., ``.", ``:", and ``$<0x0A>$"), yet frequently draw significantly higher attention scores at various locations; and \textit{Obs-(3)} attention sinks often manifest in the intermediate layers of LLMs, while the first two layers exhibit more evenly distributed attention scores, and the final layer focuses more on local information with diagonal attention patterns.


\textbf{Our answer to Q1.}
In complement to the observations made in StreamLLM, we conclude that attention sinks are found not only in the initial token but also in later tokens, particularly during the intermediate layers of LLMs.

\vspace{-0.6em}
\subsection{Q2: Will preserving attention sinks always benefit LLMs’ accuracy in different scenarios?}
\label{sec:does_help}
\vspace{-0.3em}

Considering that the newly identified attention sinks in later tokens, with their substantial attention values, divert a significant portion of attention away from other non-attention-sink tokens, it is imperative to investigate the impact of this notable diversion on the reasoning and generation capabilities of LLMs. While StreamLLM~\cite{xiao2023efficient} suggests preserving the attention sink of the initial token, it remains unclear whether preserving later attention sinks also enhances the accuracy of LLMs. Therefore, in this subsection, we delve into the impact of attention sinks on LLMs' accuracy in downstream tasks.


\textbf{Settings.} We make a heuristic attempt to verify the influence of attention sinks by decreasing the attention scores of each attention head associated with attention sinks and examining whether this can enhance the accuracy achieved by LLMs on the MMLU dataset~\cite{hendrycks2020measuring}. Taking into account the various layer-wise attention patterns discussed in Sec.~\ref{sec:does_exist}-\textit{Obs-(3)}, we only apply this operation to attention heads between the third layer and the second-to-last layer.





\zzyu{To effectively reduce the attention scores of attention sink tokens and leverage the reduced attention scores to improve the achievable performance of the target LLM by distributing them across other tokens, we propose a simple calibration technique comprising three steps:}





\begin{enumerate}
\vspace{-1em}
    \item Identify a set of attention sink tokens $\mathcal{S}_h^l = \{t\in\{1,\cdots,T\}\mid a_h^l[t] > \alpha \cdot 1/N\}$, where $\alpha=5$ by default.
    \vspace{-0.5em}
    \item Reduce the attention scores of attention sinks located in later tokens by setting 
    $\hat{A}_h^l[k, s] = A_h^l[k,s] \times \beta$ for all $s \in \mathcal{S}_h^l$ for each row $k$ in the attention map $A_h^l$, \zzyu{where $\beta$ is a hyperparameter controlling the extent to which we want to eliminate the excessive attention scores of attention sinks.}

    \vspace{-0.5em}
    \item \zzyu{To leverage the reduced attention scores, we propose to maintain the target LLM's original attention distribution to preserve token-wise relationships while slightly increasing the attention scores to enforce greater focus on the semantic information of non-attention sink tokens by setting}
    $\hat{A}_h^l[k, s] = A_h^l[k,t]+\sum_{s\in\mathcal{S}_h^l} (\hat{A}_h^l[k,s] - A_h^l[k,s]) \times \frac{A_h^l[k,t]}{\sum_{i\in\{1,\cdots,T\} - \mathcal{S}_h^l}A_h^l[k,i]}$ for all $s \notin \mathcal{S}_h^l$, which ensures that the sum of each row $k$ remains one.

    \vspace{-1em}
\end{enumerate}

\begin{figure}[bt]
    \centering
    \includegraphics[width=0.9\linewidth]{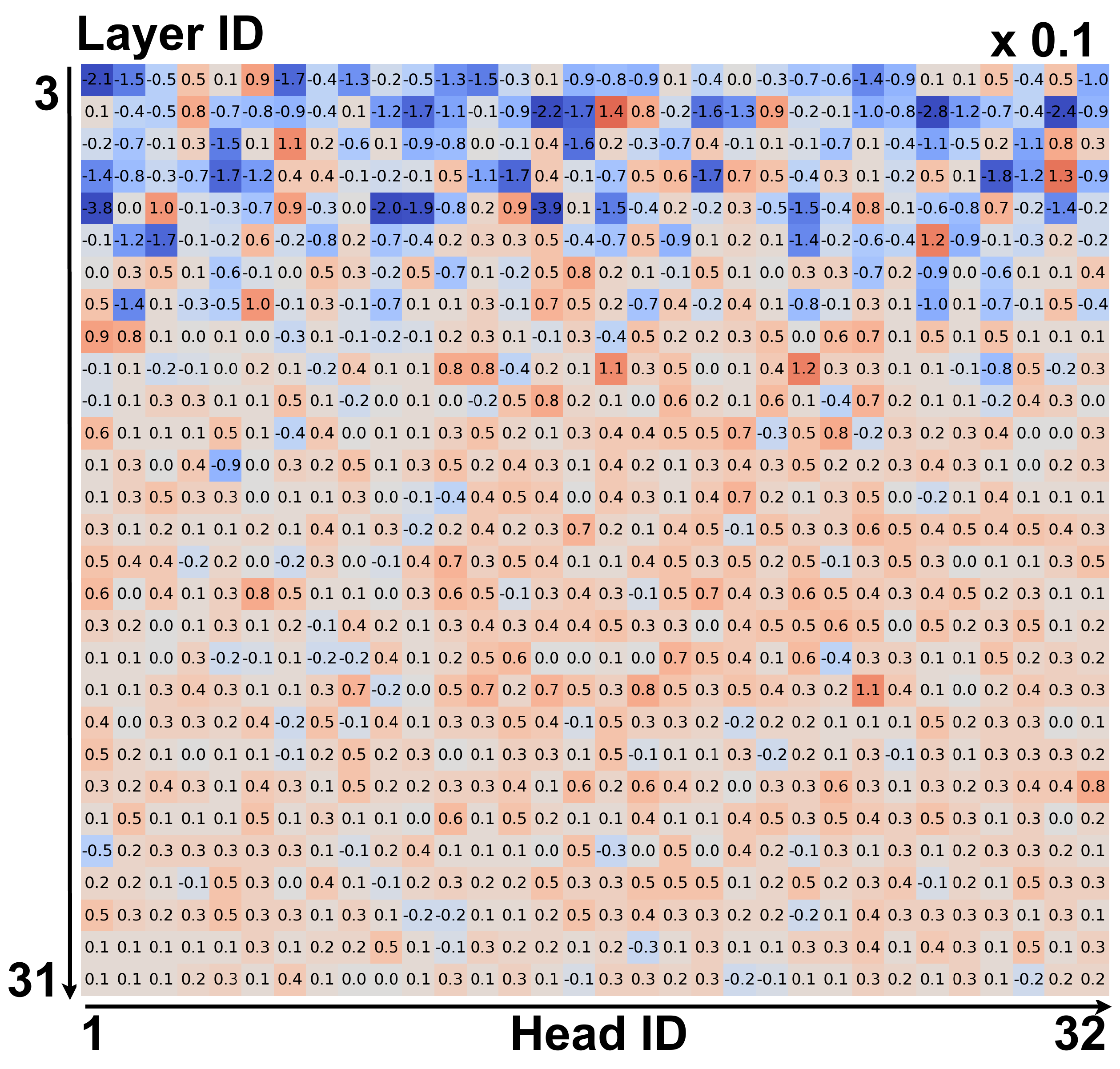}
    \vspace{-1.5em}
    \caption{Visualization of accuracy improvement in the MMLU dataset~\cite{hendrycks2020measuring} achieved by reducing the attention score of attention sinks in the middle of input sequences for each individual head separately. }
    \vspace{-1.5em}
    \label{fig:each_head_ablate}
\end{figure}

\textbf{Observations.} As demonstrated in Fig.~\ref{fig:each_head_ablate}, we can make two observations: \textit{Obs-(1)} despite the simplicity of the calibration technique we propose, in more than 76.8\% of cases, the LLM after attention calibration can achieve better accuracy compared to the vanilla inference baseline; and \textit{Obs-(2)} not all heads can benefit from the calibration, for instance, calibrating certain heads can result in an accuracy drop as significant as 0.39\%.


\textbf{Our answer to Q2.} 
In contrast to the observation made in StreamLLM~\cite{xiao2023efficient} that suggests preserving attention sinks to enhance LLMs' achievable accuracy, we highlight that \textit{not all attention sinks are beneficial for LLMs}. Specifically, for the majority of attention sinks occurring in the middle or later parts of inputs, reducing their attention scores can result in improved accuracy. We suspect this is because frequently occurring attention sinks excessively divert attention and reducing them can effectively allocate more attention to tokens with richer semantic information.

\vspace{-0.6em}
\subsection{Q3: Can we enhance LLMs' accuracy by solely manipulating attention sinks without finetuning?}
\label{sec:how_to_opt}
\vspace{-0.3em}

The observations in Sec.~\ref{sec:does_help} highlight the potential for enhancing LLMs' achievable accuracy by simply calibrating attention sinks in specific heads, even without fine-tuning. This introduces a new design parameter for improving LLMs' accuracy. However, the challenge lies in identifying the heads that require calibration, especially given that improperly reducing attention sinks in certain heads can significantly degrade LLMs' accuracy. Therefore, the remaining research question pertains to developing a technique that can automatically identify and calibrate attention sinks in the appropriate heads to enhance LLM accuracy.

\textbf{Our solution to addressing Q3.}
To enhance LLMs' accuracy without the need for finetuning, by directly optimizing attention sinks, we introduce an effective and low-cost attention calibration technique, dubbed ACT. ACT first filters out the heads that need to preserve all the corresponding attention sinks they process offline and then calibrates the attention in the remaining heads during inference.

Specifically, in the first head filtering step, we aim to determine the set of attention heads that need to preserve all the processed attention sinks, meaning that these heads should not undergo any attention calibration during inference. This filtering process can be formally described as follows:
For each task $\mathcal{T} = \{\mathcal{D}_1, \cdots, \mathcal{D}_Q\}$, consisting of $Q$ different datasets, we initially create a small held-out dataset $\mathcal{C}$ by uniformly sampling data samples from each dataset $\mathcal{D}_q \in \mathcal{T}$, ensuring that $\|\mathcal{C} \cap \mathcal{D}_q\| = M,\  \forall \mathcal{D}_q \in \mathcal{T}$ \zzyu{(i.e., each dataset $\mathcal{D}_q$ has $M$ samples in $\mathcal{C}$).}
Next, we execute the attention calibration steps as proposed in Sec.~\ref{sec:does_help}, individually on each attention head, and evaluate the resulting performance on the held-out dataset $\mathcal{C}$. Finally, we can identify a set of heads $\mathcal{H}$ that can enhance the accuracy of the target LLM after the calibration process.

In the second attention calibration step, we calibrate all $a_h^l[t]\ \forall (l,h)\in\mathcal{H}$ on the fly during inference in an input-adaptive manner, leveraging the proposed attention calibration steps in Sec.~\ref{sec:does_help} to reduce excessive attention at attention sinks.


\vspace{-0.9em}
\section{Experimental Results}
\vspace{-0.3em}

\begin{table*}[t]\centering
\caption{ACT on \textbf{domain-specific multiple choice} datasets}\label{tab:multi}
\resizebox{0.93\textwidth}{!}{
\begin{tabular}{cccccccccccc}\toprule
Model &Setting &Method & Hellaswag &ARCE &PIQA &OB &ARCC &COPA &CQA & Avg. \\\cmidrule{1-11}
\multirow{12}{*}{Llama2-7B-chat} &\multirow{3}{*}{0-shot} &Vanilla &41.65 &75.61 &63.22 &57.20 &52.17 &85.00 &59.71 &62.08 \\
& &ACT &42.70 &75.79 &66.54 &59.00 &53.85 &89.00 &59.71 &63.80 \\
& &Improv. &1.05 &0.18 &3.32 &1.80 &1.68 &4.00 &0.00 &1.72 \\\cmidrule{2-11}
&\multirow{3}{*}{1-shot} &Vanilla &30.99 &75.44 &59.25 &54.20 &53.51 &72.00 &59.54 &57.85 \\
& &ACT &31.52 &75.79 &60.55 &57.00 &54.52 &76.00 &60.04 &59.35 \\
& &Improv. &0.53 &0.35 &1.30 &2.80 &1.01 &4.00 &0.50 &1.50 \\\cmidrule{2-11}
&\multirow{3}{*}{3-shot} &Vanilla &42.46 &77.54 &65.56 &56.40 &55.52 &69.00 &62.49 &61.28 \\
& &ACT &42.93 &77.19 &66.27 &56.60 &57.19 &69.00 &63.14 &61.76 \\
& &Improv. &0.47 &-0.35 &0.71 &0.20 &1.67 &0.00 &0.65 &0.48 \\\cmidrule{2-11}
&\multirow{3}{*}{5-shot} &Vanilla &44.62 &77.02 &64.58 &59.00 &60.54 &69.00 &62.98 &62.53 \\
& &ACT &45.58 &77.72 &65.02 &59.60 &62.54 &71.00 &63.23 &63.53 \\
& &Improv. &0.96 &0.70 &0.44 &0.60 &2.00 &2.00 &0.25 &0.99 \\\cmidrule{1-11}
\multirow{12}{*}{Llama2-13B-chat} &\multirow{3}{*}{0-shot} &Vanilla &41.80 &79.82 &69.80 &63.20 &64.21 &77.00 &64.70 &65.79 \\
& &ACT &48.28 &78.77 &69.21 &63.80 &64.88 &89.00 &64.86 &68.40 \\
& &Improv. &6.48 &-1.05 &-0.59 &0.60 &0.67 &12.00 &0.16 &2.61 \\\cmidrule{2-11}
&\multirow{3}{*}{1-shot} &Vanilla &47.27 &78.07 &69.86 &62.80 &65.89 &85.00 &60.28 &67.02 \\
& &ACT &50.49 &77.19 &70.51 &62.60 &68.23 &87.00 &60.20 &68.03 \\
& &Improv. &3.22 &-0.88 &0.65 &-0.20 &2.34 &2.00 &-0.08 &1.01 \\\cmidrule{2-11}
&\multirow{3}{*}{3-shot} &Vanilla &48.26 &82.28 &69.86 &66.40 &68.90 &85.00 &66.83 &69.65 \\
& &ACT &51.64 &82.82 &70.95 &67.60 &68.90 &85.00 &66.53 &70.49 \\
& &Improv. &3.38 &0.54 &1.09 &1.20 &0.00 &0.00 &-0.30 &0.84 \\\cmidrule{2-11}
&\multirow{3}{*}{5-shot} &Vanilla &51.26 &82.81 &67.19 &69.60 &68.23 &91.00 &66.34 &70.92 \\
& &ACT &52.67 &82.11 &67.46 &68.80 &69.23 &91.00 &67.24 &71.22 \\
& &Improv. &1.41 &-0.70 &0.27 &-0.80 &1.00 &0.00 &0.90 &0.30 \\ \midrule
\multirow{3}{*}{Mistral-7B} & \multirow{3}{*}{0-shot} & Vanilla & 49.68 &  85.96 & 72.31 & 72.00 &76.25 & 87.00 & 69.21 & 73.20 \\
& & ACT & 55.82 &  87.19 & 79.22 & 74.00 &77.26 & 95.00 & 70.60 & 77.01 \\
& & Improv. & 6.14 &  1.23 & 6.91 & 2.00 & 1.01 &8.00 & 1.39 & 3.79 \\ \midrule 
\multirow{3}{*}{Llama-30B} & \multirow{3}{*}{0-shot} & Vanilla & 42.18 &  81.75 & 55.44 & 53.40 & 64.55 &82.60 & 53.32 & 61.89 \\
& & ACT & 55.44 & 83.16 & 67.46 & 62.80 &67.89 &  90.40 & 57.17 & 69.19 \\ 
& & Improv. & 13.26 & 1.41 & 12.02 & 9.40 &3.34 &  7.80 & 3.85 & 7.30 \\

\bottomrule
\end{tabular}
}
\vspace{-1em}
\end{table*}

\subsection{Evaluation settings}
\label{sec:exp_setting}
\vspace{-0.2em}

\textbf{Models, tasks, and datasets.}
\underline{Models}: We evaluate ACT on seven models, including Llama2-7B/13B-chat~\cite{touvron2023llama2}, Mistral-7B~\cite{jiang2023mistral}, Llama-30B~\cite{touvron2023llama}, GPT-J-6B~\cite{wang2021gpt}, OPT-2.7B~\cite{zhang2022opt}, and Vicuna-7B~\cite{vicuna2023}. \underline{Tasks and datasets}:  To provide a thorough evaluation of ACT, we benchmark ACT on three types of commonly used tasks with 18 different datasets, including Hellaswag~\cite{zellers2019hellaswag}, ARCE~\cite{clark2018think}, PIQA~\cite{bisk2020piqa}, OB~\cite{mihaylov2018can}, ARCC~\cite{clark2018think}, COPA~\cite{wang2019superglue}, CQA~\cite{talmor2018commonsenseqa}, and MMLU~\cite{hendrycks2020measuring} for domain-specific multiple-choice; SST2~\cite{socher2013recursive}, SST5~\cite{socher2013recursive}, MR~\cite{pang2005seeing}, AGNews~\cite{zhang2015character}, TREC~\cite{voorhees2000building}, CB~\cite{de2019commitmentbank}, and BoolQ~\cite{clark2019boolq} for text classification; and MT-Bench~\cite{zheng2024judging}, SQuADv1~\cite{rajpurkar2016squad}, and SQuADv2~\cite{rajpurkar2018know} for open-ended question answering.

\begin{table}[t]\centering
\vspace{-1em}
\caption{ACT in boosting different LLMs on the MMLU dataset}\label{tab:mmlu}
\resizebox{\linewidth}{!}{
\begin{tabular}{cccccc}\toprule
\textbf{Model} &\textbf{Llama2 7B} &\textbf{GPT-J 7B} &\textbf{Vicuna-7B} &\textbf{opt-2.7B} &Average \\\cmidrule{1-6}
zero-shot &46.50 &26.53 &48.73 &25.46 &36.80 \\\cmidrule{1-6}
zero-shot-aug &46.82 &27.62 &49.15 &25.94 &37.38 \\\cmidrule{1-6}
Improv. &0.32 &1.09 &0.42 &0.48 &0.58 \\
\bottomrule
\end{tabular}
}
\vspace{-2em}
\end{table}

\textbf{Baselines and evaluation metrics.}
\underline{Baselines}: We benchmark ACT against the vanilla inference baseline under different shot settings, including zero-shot and 1/3/5-shot in-context learning as the baseline settings. \underline{Evaluation metrics}: We use accuracy as the metric for domain-specific multiple choice and text classification tasks, and F1 score with exact match score for the open-ended question-answering task.

\begin{table*}[t]\centering
\caption{ACT on \textbf{text classification} datasets}\label{tab:cls}
\resizebox{0.91\textwidth}{!}{
\begin{tabular}{cccccccccccc}\toprule
Model &Setting &Method &SST2 &SST5 &MR &AGNews &TREC &CB &BoolQ & Avg. \\\cmidrule{1-11}
\multirow{12}{*}{Llama2-7B-chat} &\multirow{3}{*}{0-shot} &Vanilla &92.78 &47.87 &90.99 &78.17 &11.80 &69.64 &77.68 &65.07 \\
& &ACT &93.23 &47.59 &91.74 &81.76 &18.80 &69.64 &76.48 &66.36 \\
& &Improv. &0.45 &-0.28 &0.75 &3.59 &7.00 &0.00 &-1.20 &1.29 \\\cmidrule{2-11}
&\multirow{3}{*}{1-shot} &Vanilla &87.50 &44.69 &82.93 &84.87 &21.60 &76.79 &38.87 &61.11 \\
& &ACT &89.33 &45.69 &84.33 &85.62 &23.00 &78.57 &41.74 &62.49 \\
& &Improv. &1.83 &1.00 &1.40 &0.75 &1.40 &1.78 &2.87 &1.38 \\\cmidrule{2-11}
&\multirow{3}{*}{3-shot} &Vanilla &92.08 &42.62 &92.87 &75.09 &24.20 &67.86 &68.42 &64.35 \\
& &ACT &92.78 &42.51 &92.21 &76.36 &25.00 &73.21 &72.52 &65.78 \\
& &Improv. &0.70 &-0.11 &-0.66 &1.27 &0.80 &5.35 &4.10 &1.43 \\\cmidrule{2-11}
&\multirow{3}{*}{5-shot} &Vanilla &93.69 &46.87 &90.62 &85.59 &29.60 &69.64 &81.55 &67.79 \\
& &ACT &94.04 &46.62 &90.71 &86.04 &30.60 &69.64 &81.58 &68.38 \\
& &Improv. &0.35 &-0.25 &0.09 &0.45 &1.00 &0.00 &0.03 &0.58 \\\cmidrule{1-11}
\multirow{12}{*}{Llama2-13B-chat} &\multirow{3}{*}{0-shot} &Vanilla &91.86 &46.23 &90.71 &81.07 &18.00 &66.07 &80.76 &67.81 \\
& &ACT &92.20 &46.16 &90.43 &82.37 &29.00 &75.00 &81.68 &70.98 \\
& &Improv. &0.34 &-0.07 &-0.28 &1.30 &11.00 &8.93 &0.92 &3.16 \\\cmidrule{2-11}
&\multirow{3}{*}{1-shot} &Vanilla &93.69 &42.69 &86.59 &82.51 &17.20 &75.00 &64.74 &66.06 \\
& &ACT &94.27 &42.96 &87.05 &83.57 &23.40 &75.00 &65.75 &67.43 \\
& &Improv. &0.58 &0.27 &0.46 &1.06 &6.20 &0.00 &1.01 &1.37 \\\cmidrule{2-11}
&\multirow{3}{*}{3-shot} &Vanilla &92.09 &48.14 &87.52 &80.36 &15.20 &82.14 &76.87 &68.90 \\
& &ACT &92.78 &48.23 &87.62 &80.36 &22.40 &82.14 &77.29 &70.12 \\
& &Improv. &0.69 &0.09 &0.10 &0.00 &7.20 &0.00 &0.42 &1.21 \\\cmidrule{2-11}
&\multirow{3}{*}{5-shot} &Vanilla &93.23 &47.96 &92.87 &85.95 &16.40 &73.21 &81.55 &70.17 \\
& &ACT &93.46 &47.59 &93.06 &85.97 &17.20 &76.79 &81.58 &70.81 \\
& &Improv. &0.23 &-0.37 &0.19 &0.02 &0.80 &3.58 &0.03 &0.64 \\ \midrule
\multirow{3}{*}{Mistral-7B} & \multirow{3}{*}{0-shot} & Vanilla & 92.43 & 44.96	& 89.02 & 85.09 & 22.00 & 91.07 & 85.84 & 72.91 \\
& & ACT & 92.78 & 47.14 & 90.02 & 85.59 & 23.00 & 91.07 & 85.96 & 73.65 \\
& & Improv. & 0.35 & 2.18 & 1.00 & 0.50 & 1.00 & 0.00& 0.12 & 0.74 \\ \midrule
\multirow{3}{*}{Llama-30B} & \multirow{3}{*}{0-shot} & Vanilla & 80.53 & 41.78 & 81.05 & 64.37 & 28.60 & 42.86 & 65.17 & 60.25 \\
& & ACT & 85.09 & 45.59 & 85.37 & 80.53 & 29.80 & 41.07 & 65.85 & 65.37 \\
& & Improv. & 4.56 & 3.81 & 5.32 & 16.16 & 1.20 & -1.79 & 0.68 & 5.12 \\
\bottomrule
\end{tabular}
}
\vspace{-1em}
\end{table*}

\textbf{Implementation details.} We implement our ACT framework on top of PyTorch and Huggingface. For all datasets, we use the standard prompting template provided in~\cite{ouyang2022training,sanh2021multitask,hao2022structured}. Detailed prompts we used can be found in Appendix~\ref{app:prompt}. In all our experiments, unless otherwise specified, we use $\beta=0.4$ and $\|\mathcal{C}\| = 1000\times Q$, which is less than 10\% of the size of the validation datasets. During head filtering, regardless of the number of shots we evaluate, we only perform head filtering with samples using zero-shot prompts.

\vspace{-0.3em}
\subsection{Enhancing LLM accuracy with ACT}
\vspace{-0.3em}
\textbf{Domain-specific multiple choice.}
We first validate ACT on a set of commonly used domain-specific multiple-choice datasets under different settings as shown in Table~\ref{tab:multi}. ACT on average achieves an accuracy improvement of 0.30\%$\sim$7.30\% across different models and numbers of shots. The accuracy improvement can be as high as 13.26\% on a single dataset (i.e., leveraging ACT to boost Llama-30B on Hellaswag~\cite{zellers2019hellaswag} under the zero-shot setting), and applying ACT for PIQA~\cite{bisk2020piqa} under a zero-shot setting can achieve a 1.96\% higher accuracy than the vanilla inference baseline under the 5-shot in-context learning setting. Moreover, it is worth noticing that ACT has a strong ability to adapt to different evaluation settings. Specifically, although ACT only performs head filtering using samples with a zero-shot setting, ACT not only achieves average accuracy improvements of 1.72\% and 2.61\% when applied to Llama2-7B-chat and Llama2-13B-chat under the zero-shot setting, respectively, but also achieves average accuracy improvements of 1.26\%, 0.66\%, and 0.65\% when enhancing the two models under 1/3/5-shots, respectively. 

To further validate ACT's versatility and effectiveness in enhancing the performance of different types of LLMs, we apply ACT to four different kinds of LLMs including Llama2-7B-chat~\cite{touvron2023llama2}, GPT-J-6B~\cite{wang2021gpt}, OPT-2.7B~\cite{zhang2022opt}, and Vicuna-7B~\cite{vicuna2023}, and evaluate their achieved accuracy on the representative MMLU dataset~\cite{hendrycks2020measuring}. As shown in Table~\ref{tab:mmlu}, despite different model selections, ACT consistently achieves a 0.32\%$\sim$1.09\% higher accuracy over the vanilla inference baseline, proving that our proposed ACT is a general framework capable of enhancing the performance of different kinds of LLMs despite their pretraining processes, finetuning techniques, model structures, and model sizes.

\begin{table}[t]\centering
\caption{ACT on \textbf{open-ended question-answering} datasets using Llama2-chat with different sizes. Each result for SQuADv1/v2 is presented as the exact match score/F1 score.}\label{tab:openqa}
\resizebox{\linewidth}{!}{
\begin{tabular}{ccccc}\toprule
Model & Method & MT-Bench & SQuAD v1 & SQuAD v2 \\ \midrule
\multirow{3}{*}{Llama2-7B-chat} & Vanilla & 6.272 & 31.64/47.88 & 4.36/24.42 \\
& ACT & 6.406 & 41.78/64.30 & 19.52/31.30 \\ 
& Improv. & 0.134 & 10.14/16.42  & 5.16/6.88 \\ \midrule
\multirow{3}{*}{Llama2-13B-chat} & Vanilla & 6.602 & 41.77/56.00 & 19.69/27.02 \\ 
& ACT & 6.690 & 45.89/58.57 & 21.42/28.15 \\ 
& Improv. & 0.088 & 4.12/2.57  & 1.73/1.13  \\ 
\bottomrule
\end{tabular}
}
\vspace{-1.5em}
\end{table}

\textbf{Text classification.}
We further validate ACT on a set of text classification datasets under different numbers of shots and across Llama2-7B/13B-chat as shown in Table~\ref{tab:cls}. ACT shows consistent accuracy improvement over the vanilla inference baseline across different numbers of shots, datasets, and models. Under the zero-shot setting, ACT achieves average accuracy improvements of 1.29\%, 3.16\%, 0.74\%, and 5.12\% for Llama2-7B-chat, Llama2-13B-chat, Mistral-7B, and Llama-30B, respectively. Remarkably, the application of ACT leads to a peak accuracy improvement of 16.16\% when boosting the Llama-30B model on the AGNews dataset~\cite{zhang2015character} under the zero-shot condition. This set of experiments further validates the robustness of ACT in transferring between different validation scenarios. Despite its primary application of head filtering in the zero-shot scenario, ACT not only procures average accuracy improvements of 1.29\% and 3.16\% with the Llama2-7B-chat and Llama2-13B-chat models, respectively, under zero-shot conditions, but also facilitates average accuracy gains of 1.38\%, 1.43\%, and 0.58\% across 1-shot, 3-shot, and 5-shot settings for the Llama2-7B-chat. Similarly, for the Llama2-13B-chat model, ACT achieves average accuracy enhancements of 1.37\%, 1.21\%, and 0.64\% across the 1-shot, 3-shot, and 5-shot configurations, respectively.

\textbf{Open-ended question-answering.} To better validate ACT's ability to enhance LLM accuracy across different application scenarios, \zzyu{we further evaluate our proposed ACT performance on open-ended question-answering task using widely used SQuADv1~\cite{rajpurkar2016squad} and SQuADv2~\cite{rajpurkar2018know} datasets, and a more challenging multi-round conversation dataset from MT-Bench~\cite{zheng2023judging}. 
As shown in Table~\ref{tab:openqa}, ACT consistently achieves superior performance in all metrics of MT-Bench and SQuAD v1/v2 compared to vanilla inference. Specifically, ACT achieves a 0.088$\sim$0.134 higher MT-Bench score, a 1.73$\sim$10.14 higher exact match score, and a 1.13$\sim$16.42 higher F1 score over the benchmarked vanilla LLMs, respectively. It is also worth noting that an improvement of 0.088$\sim$0.134 on MT-Bench achieved by ACT is non-trivial. The difference in MT-Bench scores between Llama2-7B-chat and Llama2-13B-chat is 0.38, while the difference between Llama2-13B-chat and Llama2-70B-chat is 0.21. This suggests that applying ACT can mitigate around one-third of the difference between a smaller model and its larger counterpart.}
This proves that for the more complicated autoregressive generation task, the phenomenon that attention sinks appears in the middle part of the input sequence and draws an excessive amount of attention, sabotaging the achievable performance of LLMs still exists. Moreover, using our proposed ACT can calibrate the attention and enhance the generation quality of LLMs.

\begin{table}[t]\centering
\caption{Ablate on attention calibration methods }\label{tab:calibrate}
\resizebox{\linewidth}{!}{
\begin{tabular}{c|cccc}\toprule
Calibrate method & Temp &Inv-temp &Inv-ours &Ours \\
\midrule
Acc. & 44.89 &44.06 & 46.21 &46.82 \\
\bottomrule
\end{tabular}
}
\vspace{-1em}
\end{table}

\vspace{-0.3em}
\subsection{Ablation studies}
\label{sec:ablate}
\vspace{-0.3em}

\textbf{Ways to calibrate attention.}
We further validate whether our answers to \textbf{Q2}  in Sec.~\ref{sec:does_help} and \textbf{Q3} in Sec.~\ref{sec:how_to_opt} is correct. Specifically, we assess whether reducing the attention score at attention sinks helps improve LLM performance. To this end, we evaluate the performance of our method against three other methods: (1) Temp, which directly applies a temperature $\theta = 1.1$ to all tokens except the attention sink at the initial token; (2) Inv-temp, similar to temp but with $\theta = 1/1.1$; and (3) Inv-ours, the inverse process of our proposed method, which reduces the attention value of other tokens and redistributes it to the attention sink. As shown in Table~\ref{tab:calibrate}: (1) Our method achieves better results on MMLU compared to Temp. We attribute this improvement to our method's superior ability to preserve the original attention distribution across other tokens. (2) Inv-temp and inv-ours perform worse than temp and our method, respectively, on MMLU, indicating the importance of reducing the attention values of attention sinks in the middle part of the input.

\begin{figure}
    \centering
    \includegraphics[width=\linewidth]{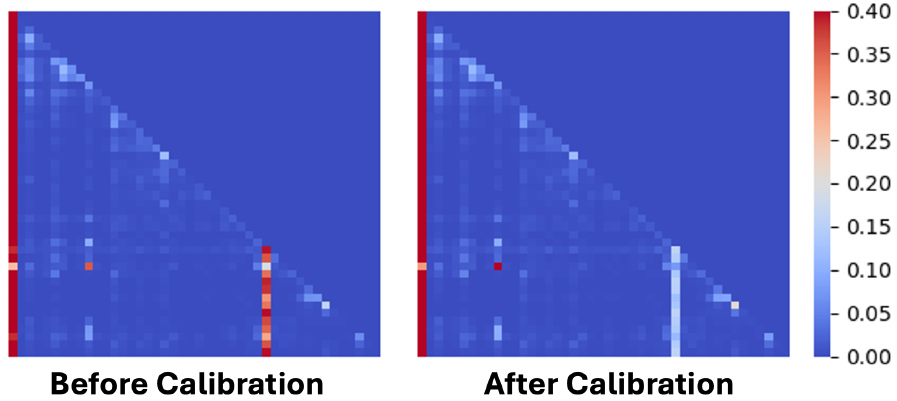}
    \vspace{-2em}
    \caption{Visualization on the model's averaged attention map before (left) and after (right) our proposed ACT.}
    \label{fig:calibrate}
    \vspace{-1.5em}
\end{figure}

\textbf{Ways to distribute the additional attention.}
After reducing the excessive attention value at attention sinks, how to distribute them across other tokens is an important question. Considering that the input of the multiple-choice dataset MMLU consists of a question and a set of choices, we evaluate three different ways to distribute the additional attention on MMLU: (1) uniform, where we uniformly distribute the additional attention across all tokens; (2) question-only, where we apply the additional attention only to the tokens corresponding to the questions; and (3) choice-only, where we apply the additional attention only to the tokens corresponding to the provided choices. As shown in Table~\ref{tab:distribute}, we observe that distributing attention to all tokens (i.e., uniform and our method) is important for preserving performance. We suspect this is because drastically changing the attention distribution across too many tokens should be avoided. 

\begin{table}[t]\centering
\caption{Ablate on how to distribute the additional attention. }\label{tab:distribute}
\resizebox{\linewidth}{!}{
\begin{tabular}{c|ccccc}\toprule
Method & Uniform &Question-only &Choices-only &Ours \\
\midrule
Acc. & 46.49 &46.10 &45.24 &46.82 \\
\bottomrule
\end{tabular}
}
\vspace{-1.6em}
\end{table}

\begin{table}[t]
    \centering   
    \caption{Ablate on $\alpha$ selection. }
    \resizebox{\linewidth}{!}{
    \begin{tabular}{c|ccccccc}\toprule
    $\alpha$ & SST2 & SST5 & MR & AGNews & TREC & CB & BoolQ  \\ \midrule
        Vanilla & 92.78 & 47.87 & 90.99 & 78.17 & 11.80 & 69.64 & 77.68 \\
        3 & 93.23 & 47.59 & 91.74 & 81.74 & 19.00 & 69.64 & 76.26 \\
        5 & 93.23 & 47.59 & 91.74 & 81.76 & 18.80 & 69.64 & 76.48 \\
        7 & 93.12 & 47.68 & 91.74 & 81.29 & 18.80 & 69.64 & 76.62 \\
        \bottomrule
    \end{tabular}
    }
    \vspace{-1.6em}
    \label{tab:alpha}
\end{table}

\textbf{$\alpha$ selection.} $\alpha$ defines the criteria of attention sink in ACT. In this paper, we empirically set $\alpha=5$ based on the visualization of the attention score distribution across different tokens, as shown in Fig.~\ref{fig:attn_distribution}. To better understand the robustness of ACT across different selections of $\alpha$, we test ACT under the zero-shot setting with Llama2-7B-chat using various values of $\alpha$. As shown in Table~\ref{tab:alpha}, despite different selections of $\alpha$, ACT consistently achieves similar performance with a steady 1.24\%$\sim$1.29\% higher average accuracy than the vanilla Llama2-7B-chat baseline. This demonstrates that the attention sinks identified in our work have distinct values compared to other non-attention sink tokens, and thus, the selection of $\alpha$ plays a minor role in the performance of ACT.

\textbf{$\beta$ selection.}
$\beta$ determines how drastically we want to reduce the attention sinks that occur in the middle of the input. In this paper, we set $\beta=0.4$, but we want to explore the impact of $\beta$ selection on the final achieved accuracy on MMLU with Llama2-7B-chat. As shown in Table~\ref{tab:beta}, despite different selections of $\beta$ result in varied accuracies, they all achieve better accuracies than the vanilla inference baseline, showing ACT is robust to different hyperparameter selections. 

\begin{table}[t]\centering
\caption{Ablate on $\beta$ selection.}\label{tab:beta}
\resizebox{0.9\linewidth}{!}{
\begin{tabular}{c|ccccccc}\toprule
$\beta$ &Vanilla &0.7 &0.5 & 0.4 (Ours) &0.3 &0.1 \\
\midrule
Acc. &46.50 &46.77 &46.81 &46.82 & 46.79 &46.65 \\
\bottomrule
\end{tabular}
}
\vspace{-1.6em}
\end{table}

\begin{table}[!htb]\centering
\caption{Ablate on $M$ selection.}\label{tab:dataset_size}
\resizebox{0.9\linewidth}{!}{
\begin{tabular}{c|cccccc}\toprule
 $M$ & Vanilla  &300 &600 &1000 & All \\
 \midrule
Acc. &46.50 &46.50 &46.56 &46.82 &46.91 \\
\bottomrule
\end{tabular}
}
\vspace{-1.6em}
\end{table}

\textbf{Size of $\mathcal{C}$.}
We ablate the appropriate size of $\|\mathcal{C}\|$, which controls nearly the only source of overhead in ACT. We ablate different selections of $\|\mathcal{C}\|$ by sampling different numbers of samples in each $\mathcal{D}_q \in \mathcal{Q}$ (i.e., $\|\mathcal{C}\| / Q$) and evaluating their achieved accuracy on the MMLU dataset using Llama2-7B-chat. As shown in Table~\ref{tab:dataset_size}, a larger $\mathcal{C}$ helps with ACT's performance, but when $\|\mathcal{C}\| / Q$ scales up to around 1000 (e.g., more than 10 times smaller than the validation dataset), the further performance improvement is marginal.

\begin{table}[!t]
    \centering
    \caption{Ablate on the performance of ACT when only calibrating on a subset of the selected attention heads.}
    \resizebox{\linewidth}{!}{
    \begin{tabular}{c|ccccc}\toprule
        Subset size &  SST2 & AGNews & PIQA & ARCC & Avg.\\ \midrule
        0\% (Vanilla) & 92.78 & 78.17 & 63.22 & 52.10 & 71.57 \\ 
        40\% & 92.78 & 80.16 & 66.92 & 53.51 & 73.34 \\ 
        60\% & 92.89& 81.12 &65.34 & 52.17 & 72.88\\
        80\% & 93.23 & 81.08 & 66.63 & 52.84 & 73.44 \\ 
        100\% (ACT) & 93.23 & 81.76 & 66.54 & 53.85 & 73.84 \\
        \bottomrule
    \end{tabular}
    }
    \vspace{-1.2em}
    \label{tab:num_head}
\end{table}

\textbf{Number of heads to calibrate.} To verify whether the performance improvement achieved by calibrating each individual attention head as in Fig.~\ref{fig:each_head_ablate} can be accumulated, we validate ACT's performance when calibrating on subsets of $\mathcal{H}$ of different sizes. As shown in Table~\ref{tab:num_head}, the achieved performance of attention calibration gradually increases as the size of the subsets increases, validating that the effectiveness of calibrating each attention head in $\mathcal{H}$ can be accumulated and that calibrating all heads in $\mathcal{H}$ leads to optimal performance.

\vspace{-0.3em}
\subsection{Attention map visualization before and after ACT}
\vspace{-0.3em}

To better understand the role of our proposed ACT in reducing the excessive attention at attention sinks in the middle of inputs, we further visualize the attention map of Llama2-7B-chat before and after performing ACT with the same input sample. As shown in Fig.~\ref{fig:calibrate}, after performing ACT, the original attention sink that occurs in the middle of the input sequence is almost eliminated, while the attention distribution of other tokens remains the same. 
\vspace{-0.6em}
\section{Conclusion}
\label{sec:conclusion}
\vspace{-0.3em}
In this paper, we conduct comprehensive visualizations of the attention distributions in LLMs during inference across various inputs and tasks. Based on these visualizations, for the first time, we discover that (1) attention sinks occur not only at the start of sequences but also within later tokens of the input, and (2) not all attention sinks have a positive impact on the achievable accuracy of LLMs. Building upon our findings, we propose a training-free technique, dubbed ACT, that automatically optimizes the attention distributions on the fly during inference in an input-adaptive manner. Extensive experiments validate that ACT consistently enhances the accuracy of various LLMs across different applications. 

\newpage
\section*{Impact Statement}
The recent advancements in LLMs have triggered various application scenarios that require an affordable LLM with superior performance to serve as a backbone. This calls for (1) LLMs with better performance under comparable computation costs and (2) a better understanding of the behavior of LLMs, facilitating a trustworthy generation process. In this paper, we cater to both of the aforementioned calls.

For (1), our proposed ACT can improve the performance of LLMs on downstream tasks not only in a training-free manner but also with almost no additional inference cost. The proposed ACT leverages the design knob on attention manipulation, which is also orthogonal to most techniques improving the performance of LLMs, such as in-context learning, prompting, and finetuning, making ACT a generally applicable technique.

For (2), we have conducted comprehensive visualization and analysis of the attention generated by LLMs during inference with different inputs from various tasks. Moreover, to the best of our knowledge, we are the first to discover that attention sinks manifest not only in the initial token but also in subsequent tokens throughout the input context. This observation deepens our understanding of the intrinsic mechanism of LLMs and thus can potentially facilitate the trustworthy generation process.

\section*{Acknowledgements}
This work is supported by the National Science Foundation
(NSF) through the CCRI funding (Award number: 2016727) and CoCoSys, one of seven centers in JUMP 2.0, a Semiconductor Research Corporation (SRC) program sponsored by DARPA.

\nocite{langley00}

\bibliography{refer}
\bibliographystyle{icml2024}

\newpage
\appendix
\onecolumn



\section{Histogram on the position of attention sinks}
\begin{figure*}[h]
    \centering
    \includegraphics[width=\textwidth]{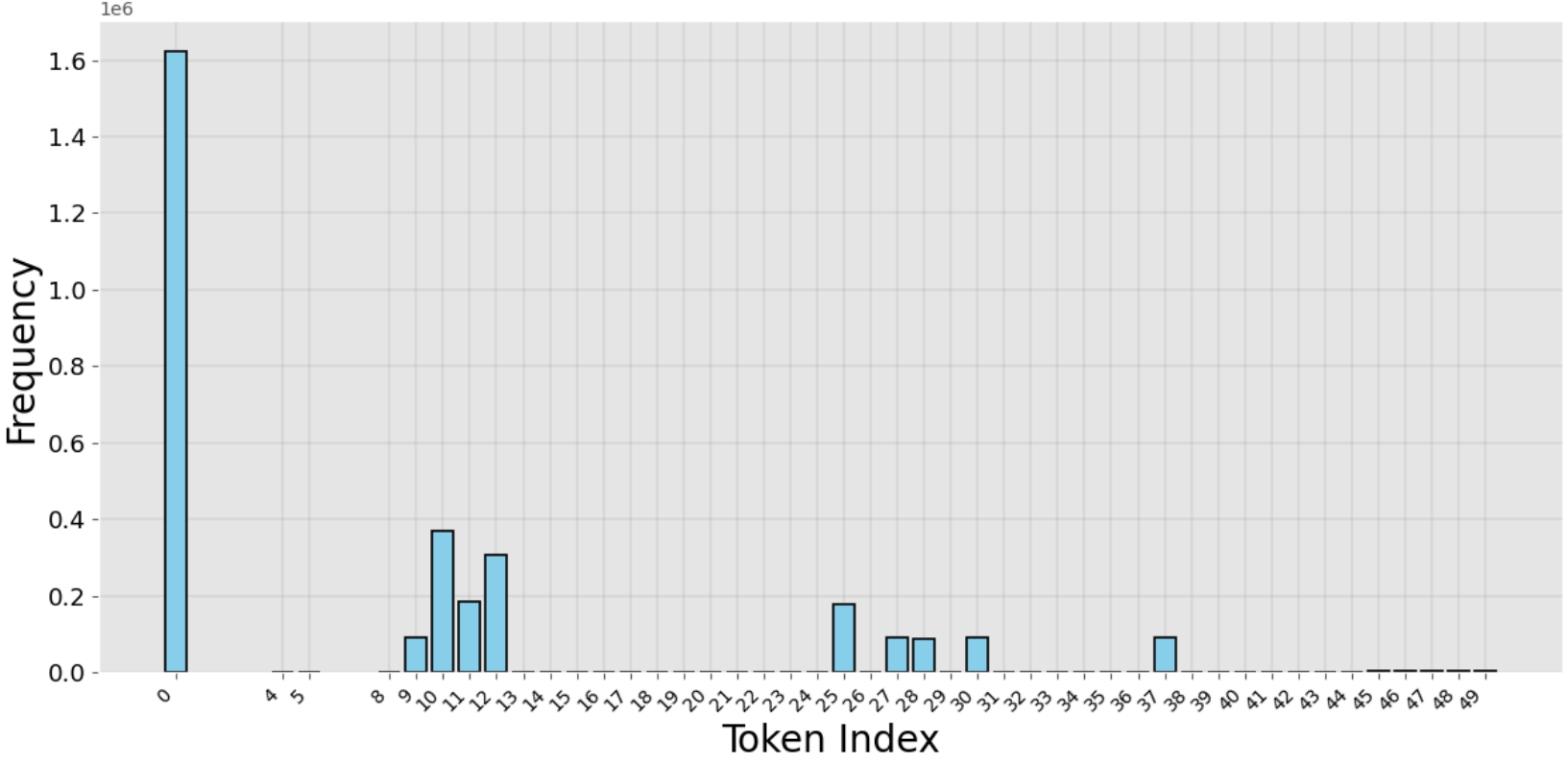}
    \vspace{-2em}
    \caption{Histogram of the positions of attention sinks throughout all 17 datasets used in our paper. }
    \label{fig:histogram}
\end{figure*}

To better understand the attention sink distribution, we profile all of the attention sink that occurred during inference with Llama2-7B-chat on all 17 datasets mentioned in Sec.~\ref{sec:exp_setting}. As shown in Fig.~\ref{fig:histogram}, despite the attention sink at the initial token occurring the most frequently, there are many other positions that are prone to have attention sink, further proving the wide existence of attention sink phenomenon throughout the input sequence. 


\section{Prompts used for each dataset}
\label{app:prompt}
Here, we list all the prompts we used in this paper on different datasets: 

For multiple choice task (i.e., on hellaswag, ARCE, PIQA, OB, ARCC, COPA, CQA datasets), we use the following prompt: 
\begin{itemize}
    \item  "Complete the following sentence with an appropriate ending.
    
    $<$Question$>$
    
    $<$choice 1$>$

$<$choice 2$>$

$<$choice 3$>$

…

Answer:"
\end{itemize}

For MMLU datasets, we use the following prompt: 
\begin{itemize}
    \item  "The following are multiple choice questions (with answers) about $<$subject$>$.

    $<$Question$>$
    
    $<$choice 1$>$

$<$choice 2$>$

$<$choice 3$>$

…

Answer:"
\end{itemize}

For text classification, we use different prompts for different datasets. 
\begin{itemize}
    \item SST2:
    \begin{itemize}
        \item "Classify the sentiment of the user's message into one of the following categories:'positive' or 'negative'.

        -
        
        Sentence: $<$sentence$>$ 
        
        Sentiment: "
    \end{itemize}

    \item SST5: 
    \begin{itemize}
        \item  "Classify the sentiment of the user's message into one of the following categories:'terrible', 'negative', 'neutral', 'positive', or 'great'.
        
        -

        Sentence: $<$sentence$>$  
        
        Sentiment: "
    \end{itemize}

    \item MR:
    \begin{itemize}
        \item ``Classify the sentiment of the movie's review into one of the following categories:'positive' or 'negative'.
        
        -

        Review:  $<$sentence$>$  
        
        Sentiment: "
    \end{itemize}
    \item AGNews: 
    \begin{itemize}
        \item  "Classify the news articles into the categories of 'World', 'Sports', 'Business', or 'Technology'.
        
        -

        Article: $<$sentence$>$  
        
        Category: "
    \end{itemize}
    \item TREC:
    \begin{itemize}
        \item "Classify the given questions into the following categories of 'Description', 'Entity', 'Expression', 'Person', 'Number', or 'Location'.
        
        -

        Question:  $<$sentence$>$  
        
        Type: "
    \end{itemize}
    \item CB:
    \begin{itemize}
        \item "Read the following paragraph and determine if the hypothesis is true.
        
        -
         
         Premise: $<$premise$>$  Hypothesis: $<$hypothesis$>$. Answer: "
    \end{itemize}
    \item BoolQ:
    \begin{itemize}
        \item "Read the text and answer the question by True or False.
        
        -
         
         Text: $<$passage$>$ Question: $<$question$>$? 
         
         Answer: "
    \end{itemize}
\end{itemize}

For open-ended question answering (i.e., SQuADv1/v2), we use the following prompt: 
\begin{itemize}
    \item Answer question using information in the preceding background paragraph.
If there is not enough information provided, answer with “Not in background.”

Title: [title]

Background: [background]

Q: [first question]

A: [first answer]

Q: [final question]

A: [completion]
\end{itemize}


\section{More visualizations on attention maps}
\begin{figure}[h]
    \centering
    \includegraphics[width=0.8\textwidth]{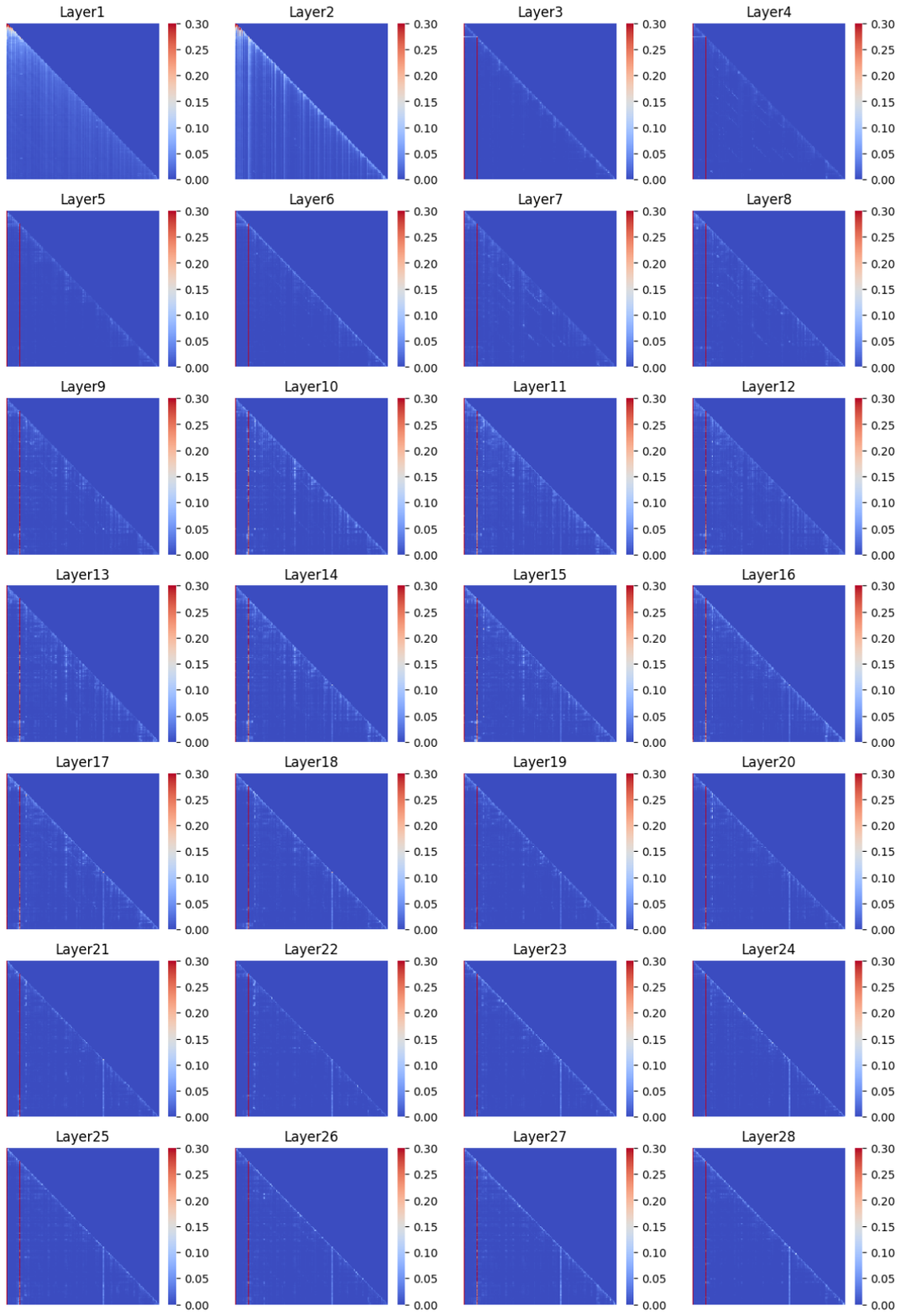}
    \caption{Visualization on the attention map of each layer in \textbf{Llama2-7B-chat} model when processing the following input sample: 'Read the text and answer the question by True or False.$\backslash$n$\backslash$nText: Riverdale (2017 TV series) -- The series debuted on January 26, 2017 to positive reviews. A 22-episode second season premiered on October 11, 2017, and concluded on May 16, 2018. On April 2, 2018, The CW renewed the series for a third season, which is set to premiere October 10, 2018. Question: is there going to be any more episodes of riverdale? $\backslash$n Answer: '}
    \label{fig:llama}
\end{figure}

\begin{figure}[h]
    \centering
    \includegraphics[width=0.8\textwidth]{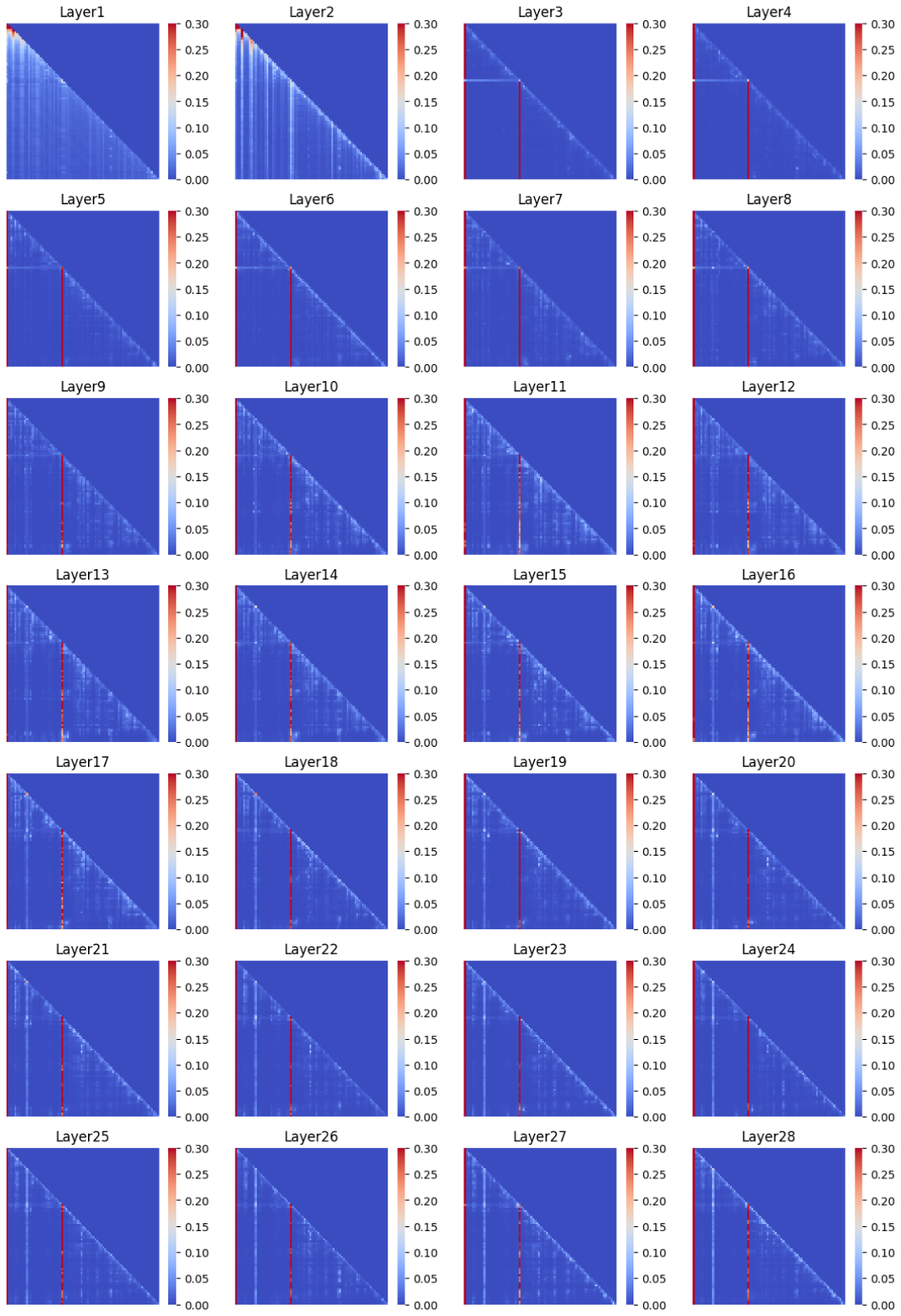}
    \caption{Visualization on the attention map of each layer in \textbf{Vicuna-7B} model when processing the following input sample: "Classify the sentiment polarity of the movie's review into one of the following categories: 'subjective' or 'object'.$\backslash$n$\backslash$nInput: when all seems hopeless, ted gets some guidance from his good friend meg that turns the situation around: `` don't scam on her, listen to her, be sincere. `` $\backslash$nType: "}
    \label{fig:vicuna}
\end{figure}

\begin{figure}[h]
    \centering
    \includegraphics[width=0.8\textwidth]{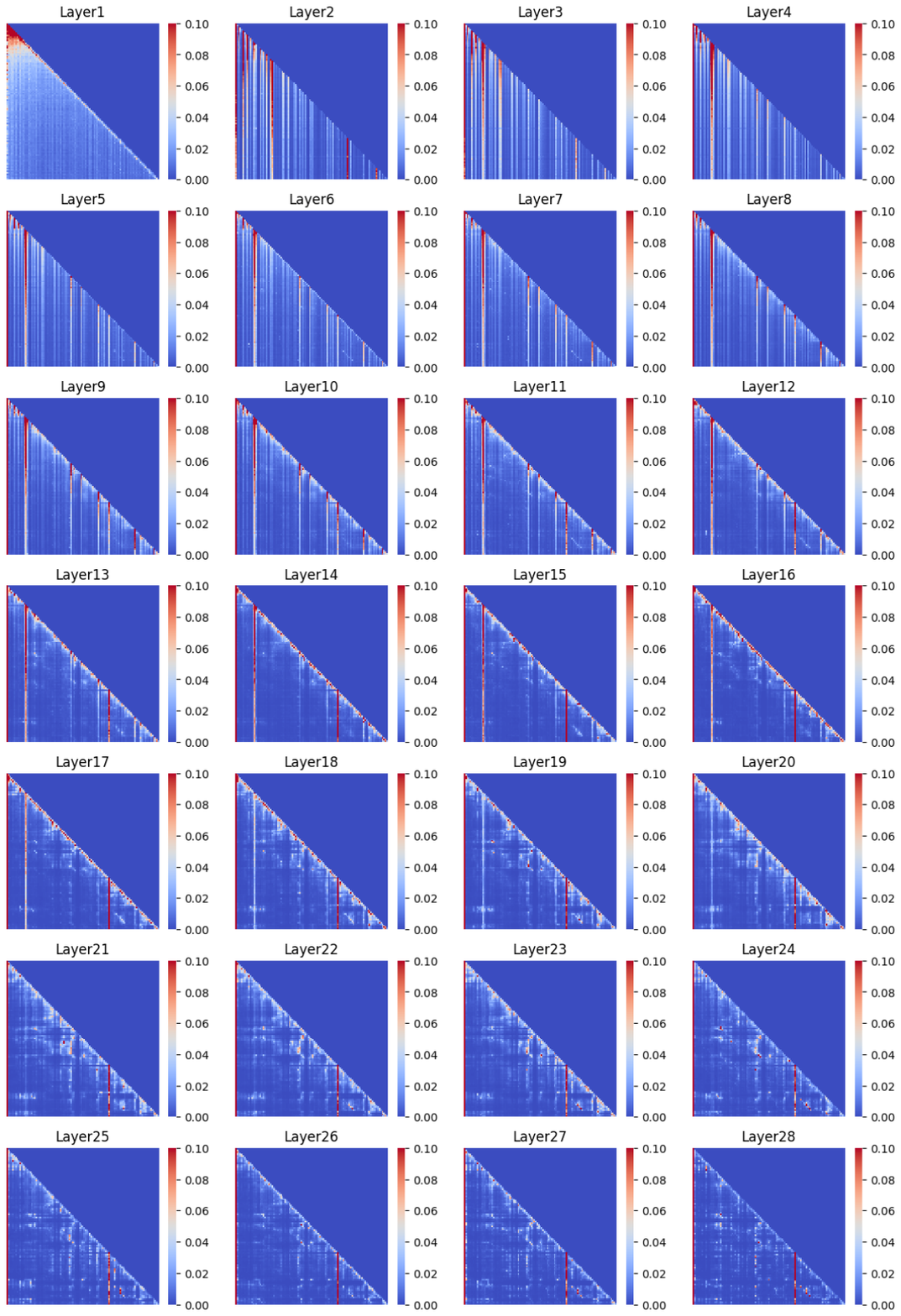}
    \caption{Visualization on the attention map of each layer in \textbf{OPT-2.7B} model when processing the following input sample: ""Read the following paragraph and determine if the hypothesis is true. $\backslash$n $\backslash$n Premise: A: Oh, oh yeah, and every time you see one hit on the side of the road you say is that my cat. B: Uh-huh. A: And you go crazy thinking it might be yours. B: Right, well I didn't realize my husband was such a sucker for animals until I brought one home one night. Hypothesis: her husband was such a sucker for animals. Answer: " "}
    \label{fig:opt}
\end{figure}

We conduct more visualization on different LLMs as shown in Fig.~\ref{fig:llama}, Fig.~\ref{fig:vicuna}, and Fig.~\ref{fig:opt} for Llama2-7B-chat, Vicuna-7B, and OPT-2.7B, respectively.

\label{app:avg_attn_vis}

\end{document}